\documentclass{article} 
\PassOptionsToPackage{numbers}{natbib}

\usepackage[final]{nips_2016_preprint}
\usepackage{times}
\usepackage{graphicx} 
\usepackage{caption}
\usepackage{subcaption}

\usepackage{comment}

\usepackage{algorithm}
\usepackage{algorithmic}
\usepackage{mathrsfs}

\usepackage{url}
\usepackage{amssymb}
\usepackage{amsmath}
\usepackage{amsthm}
\usepackage{chngcntr}

\usepackage{color}
\usepackage{lscape}
\usepackage[figuresleft]{rotating}
\usepackage{sidecap}
\usepackage{booktabs, topcapt}

\newcommand{\shuyang}[1]{{\color{black} #1}}

\newcommand{\be}{\begin{eqnarray} \begin{aligned}}
\newcommand{\ee}{\end{aligned} \end{eqnarray} }
\newcommand{\benn}{\begin{eqnarray*} \begin{aligned}}
\newcommand{\eenn}{\end{aligned} \end{eqnarray*} }
\newcommand{\vx}{{\mathbf x}}
\newcommand{\vy}{{\mathbf y}}

\makeatletter
\def\thm@space@setup{%
  \thm@preskip=\parskip \thm@postskip=0pt
}
\makeatother

\newtheorem{theorem}{Theorem}[section]
\newtheorem*{theorem*}{Theorem}

\title{Variational Information Maximization for \\
Feature Selection}
\author{
Shuyang Gao \qquad \qquad Greg Ver Steeg \qquad \qquad  Aram Galstyan \\
University of Southern California, Information Sciences Institute \\
 \texttt{gaos@usc.edu}, \texttt{gregv@isi.edu}, \texttt{galstyan@isi.edu}
}
\begin{document}

\maketitle

\begin{abstract}
Feature selection is one of the most fundamental problems in machine learning. An extensive body of work on information-theoretic feature selection exists which is based on maximizing mutual information between subsets of features and class labels. Practical methods are forced to rely on approximations due to the difficulty of estimating mutual information. We demonstrate that approximations made by existing methods are based on unrealistic assumptions. We formulate a more flexible and general class of assumptions based on variational distributions and use them to tractably generate lower bounds for mutual information. These bounds define a novel information-theoretic framework for feature selection, which we prove to be optimal under tree graphical models with proper choice of variational distributions. Our experiments demonstrate that the proposed method strongly outperforms existing information-theoretic feature selection approaches.

\end{abstract}
\section{Introduction}

Feature selection is one of the fundamental problems in machine learning research~\cite{dash1997feature, liu2012feature}. Many problems include a large number of features that are either \textit{irrelevant} or \textit{redundant} for the task at hand. In these cases, it is often advantageous to pick a smaller subset of features to avoid over-fitting, to speed up computation, or simply to improve the interpretability of the results. 

Feature selection approaches are usually categorized into three groups: \textit{wrapper}, \textit{embedded} and \textit{filter}~\cite{kohavi1997wrappers, guyon2003introduction, brown2012conditional}. The first two methods, \textit{wrapper} and \textit{embedded}, are considered \textit{classifier-dependent}, i.e., the selection of features somehow depends on the classifier being used. \textit{Filter} methods, on the other hand, are \textit{classifier-independent} and define a scoring function between features and labels in the selection process.

Because \textit{filter} methods may be employed in conjunction with a wide variety of classifiers, it is important that the scoring function of these methods is as general as possible. Since mutual information (MI) is a general measure of dependence with several unique properties~\cite{cover2012elements}, many MI-based scoring functions have been proposed as \textit{filter} methods~\cite{battiti1994using, yang1999data, fleuret2004fast, peng2005feature, rodriguez2010quadratic, nguyen2014effective}; see ~\cite{brown2012conditional} for an exhaustive list. 

Owing to the difficulty of estimating mutual information in high dimensions, most existing MI-based feature selection methods are based on various low-order approximations for mutual information. While those approximations have been successful in certain applications, they are heuristic in nature and lack theoretical guarantees. In fact, as we demonstrate below (Sec.~\ref{sec:background_limit}), a large family of approximate methods are based on two assumptions that are mutually inconsistent. 

To address the above shortcomings, in this paper we introduce a novel feature selection method based on variational lower bound on mutual information; a similar bound was previously studied within the Infomax learning framework~\cite{agakov2004algorithm}. We show that instead of maximizing the mutual information, which is intractable in high dimensions (hence the introduction of many heuristics), we can maximize a lower bound on the MI with the proper choice of tractable variational distributions. We use this lower bound to define an objective function and derive a forward feature selection algorithm. 

We provide a rigorous proof that the forward feature selection is optimal under tree graphical models by choosing an appropriate variational distribution. This is in contrast with previous information-theoretic feature selection methods, which lack any performance guarantees. We also conduct empirical validation on various datasets and demonstrate that the proposed approach outperforms state-of-the-art information-theoretic feature selection methods.




In Sec.~\ref{sec:background} we introduce general MI-based feature selection methods and discuss their limitations. Sec.~\ref{sec:method} introduces the variational lower bound on mutual information and proposes two specific variational distributions. In Sec.~\ref{sec:exp}, we report results from our experiments, and compare the proposed approach with existing methods.

\section{Information-Theoretic Feature Selection Background}\label{sec:background}

\subsection{Mutual Information-Based Feature Selection}
Consider a supervised learning scenario where $\vx=\{\vx_1,\vx_2,...,\vx_D\}$ is a $D$-dimensional input feature vector, and $\vy$ is the output label. In \textit{filter} methods, the mutual information-based feature selection task is to select $T$ features $\vx_{S^*} = \{\vx_{f_1}, \vx_{f_2},...,\vx_{f_T}\}$ such that the mutual information between $\vx_{S^*}$ and $\vy$ is maximized. Formally, 
\be \label{eq:fs_obj}
{{{S^*}}} = \mathop {\arg \max }\limits_S I\left( {{{\bf{x}}_S}:{\bf{y}}} \right) \ \ \ s.t. \ |S| = T
\ee
where $I(\cdot)$ denotes the mutual information~\cite{cover2012elements}.

{\bf Forward Sequential Feature Selection \quad} 
Maximizing the objective function in Eq.~\ref{eq:fs_obj} is generally NP-hard. Many MI-based feature selection methods adopt a greedy method, where features are selected incrementally, one feature at a time. Let $S^{t-1}=\{\vx_{f_1}, \vx_{f_2},...,\vx_{f_{t-1}}\}$ be the selected feature set after time step $t-1$. According to the greedy method, the next feature $f_{t}$ at step $t$ is selected such that
\be\label{eq:fw_sel}
{f_{t}} = \mathop {\arg \max }\limits_{i \notin {S^{t-1}}} I\left( {{{\bf{x}}_{ {S^{t-1}} \cup i}}:{\bf{y}}} \right)
\ee
where ${{\bf{x}}_{ {S^{t-1}} \cup i}}$ denotes $\vx$'s projection into the feature space $S^{t-1} \cup i$.
As shown in~\cite{brown2012conditional}, the mutual information term in Eq.~\ref{eq:fw_sel} can be decomposed as:
\be\label{eq:mi_decompose}
I\left( {{{\bf{x}}_{{S^{t - 1}} \cup i}}:{\bf{y}}} \right) &= I\left( {{{\bf{x}}_{{S^{t - 1}}}}:{\bf{y}}} \right) + I\left( {{{\bf{x}}_i}:{\bf{y}}|{{\bf{x}}_{{S^{t - 1}}}}} \right)\\
 &= I\left( {{{\bf{x}}_{{S^{t - 1}}}}:{\bf{y}}} \right) + I\left( {{{\bf{x}}_i}:{\bf{y}}} \right) - I\left( {{{\bf{x}}_i}:{{\bf{x}}_{{S^{t - 1}}}}} \right) + I\left( {{{\bf{x}}_i}:{{\bf{x}}_{{S^{t - 1}}}}|{\bf{y}}} \right)\\
 &= I\left( {{{\bf{x}}_{{S^{t - 1}}}}:{\bf{y}}} \right) + I\left( {{{\bf{x}}_i}:{\bf{y}}} \right) \\
 &~~~~- \left( {H\left( {{{\bf{x}}_{{S^{t - 1}}}}} \right) - H\left( {{{\bf{x}}_{{S^{t - 1}}}}|{{\bf{x}}_i}} \right)} \right) + \left( {H\left( {{{\bf{x}}_{{S^{t - 1}}}}|{\bf{y}}} \right) - H\left( {{{\bf{x}}_{{S^{t - 1}}}}|{{\bf{x}}_i},{\bf{y}}} \right)} \right)\\
\ee
where $H(\cdot)$ denotes the entropy~\cite{cover2012elements}. Omitting the terms that do not depend on $\vx_i$ in Eq.~\ref{eq:mi_decompose}, we can rewrite Eq.~\ref{eq:fw_sel} as follows:
\be\label{eq:fw_sel_new}
{f_{t}} = \mathop {\arg \max }\limits_{i \notin {S^{t-1}}} I\left( {{{\bf{x}}_i}:{\bf{y}}} \right) + H\left( {{{\bf{x}}_{{S^{t - 1}}}}|{{\bf{x}}_i}} \right) - H\left( {{{\bf{x}}_{{S^{t - 1}}}}|{{\bf{x}}_i},{\bf{y}}} \right)
\ee
The greedy learning algorithm has been analyzed in~\cite{das2011submodular}. 

\subsection{Limitations of Previous MI-Based Feature Selection Methods}
\label{sec:background_limit}

Estimating high-dimensional information-theoretic quantities is a difficult task. Therefore most MI-based feature selection methods propose low-order approximation to $H\left( {{{\bf{x}}_{{S^{t - 1}}}}|{{\bf{x}}_i}} \right)$ and $H\left( {{{\bf{x}}_{{S^{t - 1}}}}|{{\bf{x}}_i},{\bf{y}}} \right)$ in Eq.~\ref{eq:fw_sel_new}. A general family of methods rely on the following approximations~\cite{brown2012conditional}:
\be \label{eq:mrmr}
&~H\left( {{{\bf{x}}_{{S^{t - 1}}}}|{{\bf{x}}_i}} \right) \approx \sum\limits_{k = 1}^{t - 1} {H\left( {{{\bf{x}}_{{f_k}}}|{{\bf{x}}_i}} \right)} \\
&~H\left( {{{\bf{x}}_{{S^{t - 1}}}}|{{\bf{x}}_i},{\bf{y}}} \right) \approx \sum\limits_{k = 1}^{t - 1} {H\left( {{{\bf{x}}_{{f_k}}}|{{\bf{x}}_i},{\bf{y}}} \right)} 
\ee

The approximations in Eq.~\ref{eq:mrmr} become exact under the following two assumptions~\cite{brown2012conditional}: 

\textit{Assumption 1. (Feature Independence Assumption)} $p\left( {{{\bf{x}}_{{S^{t - 1}}}}|{{\bf{x}}_i}} \right) = \prod\limits_{k = 1}^{t - 1} {p\left( {{{\bf{x}}_{{f_k}}}|{{\bf{x}}_i}} \right)} $ \\
\textit{Assumption 2. (Class-Conditioned Independence Assumption)} ~~~$p\left( {{{\bf{x}}_{{S^{t - 1}}}}|{{\bf{x}}_i},{\bf{y}}} \right) = \prod\limits_{k = 1}^{t - 1} {p\left( {{{\bf{x}}_{{f_k}}}|{{\bf{x}}_i},{\bf{y}}} \right)}$
\textit{Assumption 1} and \textit{Assumption 2} mean that the selected features are independent and class-conditionally independent, respectively, given the unselected feature $\vx_i$ under consideration. 
\begin{figure}[htbp] 
   \centering 
   \begin{subfigure}[t]{0.3\textwidth}
   \centering
   \includegraphics[height=1.0in]{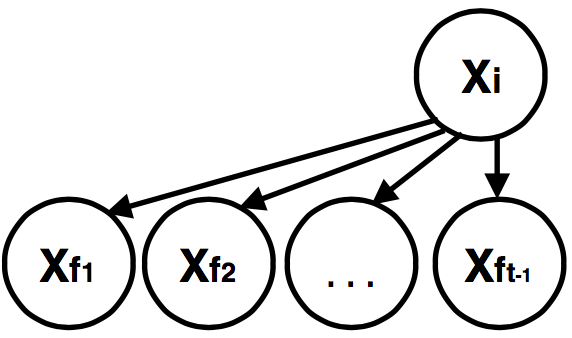}
   \caption*{\textit{Assumption 1}}
   \end{subfigure}\quad\quad
   \begin{subfigure}[t]{0.3\textwidth}
   \centering
   \includegraphics[height=1.0in]{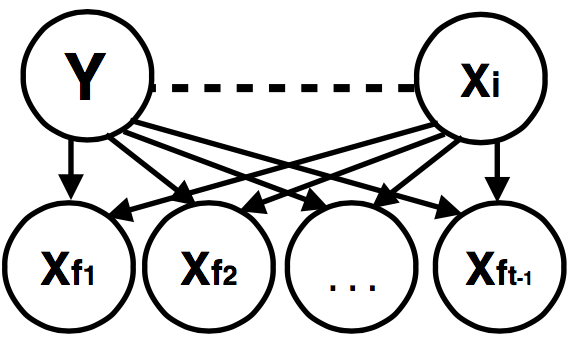}
   \caption*{\textit{Assumption 2}}
   \end{subfigure}
   \begin{subfigure}[t]{0.3\textwidth}
   \centering
   \includegraphics[height=1.0in]{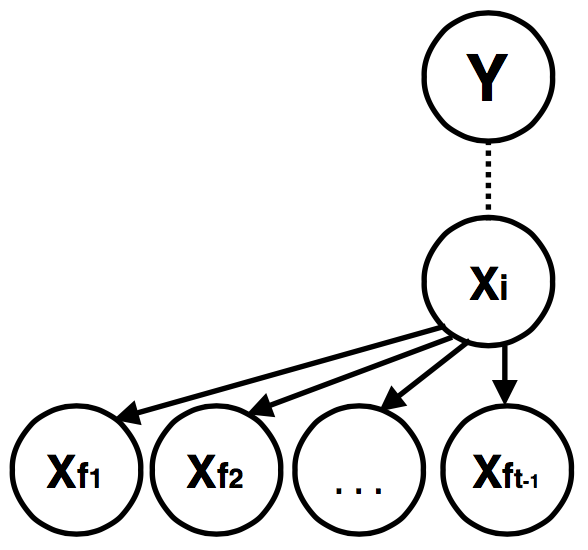}
   \caption*{Satisfying both \textit{Assumption 1} and \textit{Assumption 2}}
   \end{subfigure}
   \caption{The first two graphical models show the assumptions of traditional MI-based feature selection methods. The third graphical model shows a scenario when both \textit{Assumption 1} and \textit{Assumption 2} are true. Dashed line indicates there may or may not be a correlation between two variables.}
    \label{fig:limit} 
\end{figure}

We now demonstrate that the two assumptions cannot be valid simultaneously unless the data has a very specific (and unrealistic) structure. Indeed, consider the graphical models consistent with either assumption, as illustrated in Fig.~\ref{fig:limit}. If \textit{Assumption 1} holds true, then $\vx_i$ is the only common cause of the previously selected features  $S^{t-1}=\{\vx_{f_{1}}, \vx_{f_2}, ...,\vx_{f_{t-1}}\}$, so that those features become independent when conditioned on $\vx_i$. On the other hand, if \textit{Assumption 2} holds, then the features depend both on $\vx_i$ and class label $\vy$; therefore, generally speaking, distribution over those features does not factorize by solely conditioning on $\vx_i$---there will be remnant dependencies due to $\vy$. Thus, if \textit{Assumption 2} is true, then \textit{Assumption 1} cannot be true in general, unless the data is generated according to a very specific model shown in the rightmost model in Fig.~\ref{fig:limit}. Note, however, that in this case, $\vx_i$ becomes the most important feature because $I(\vx_i:\vy) > I(\vx_{S^{t-1}}:\vy)$; then we should have selected $\vx_i$ at the very first step, contradicting the feature selection process.
 
As we mentioned above, most existing methods implicitly or explicitly adopt both assumptions or their stronger versions as shown in~\cite{brown2012conditional}, including mutual information maximization (MIM)~\cite{lewis1992feature}, joint mutual information (JMI)~\cite{yang1999data}, conditional mutual information maximization (CMIM)~\cite{fleuret2004fast}, maximum relevance minimum redundancy (mRMR)~\cite{peng2005feature}, conditional infomax feature extraction (CIFE)~\cite{lin2006conditional}, etc. Approaches based on global optimization of mutual information, such as quadratic programming feature selection ($\mathcal{QPFS}$)~\cite{rodriguez2010quadratic} and state-of-the-art conditional mutual information-based spectral method ($\mathcal {SPEC_{CMI}}$)~\cite{nguyen2014effective}, are derived from the previous greedy methods and therefore also implicitly rely on those two assumptions. 



In the next section we address these issues by introducing a novel information-theoretic framework for feature selection. Instead of estimating mutual information and making mutually inconsistent assumptions, our framework formulates a tractable variational lower bound on mutual information, which allows a more flexible and general class of assumptions via appropriate choices of variational distributions.

\section{Method} \label{sec:method}
\subsection{Variational Mutual Information Lower Bound}
Let $p(\vx,\vy)$ be the joint distribution of input ($\vx$) and output ($\vy$) variables. Barber \& Agkov~\cite{agakov2004algorithm} derived the following lower bound for mutual information $I(\vx:\vy)$ by using the non-negativity of KL-divergence, i.e., $\sum\nolimits_{\bf{x}} {p\left( {{\bf{x}}|{\bf{y}}} \right)\log \frac{{p\left( {{\bf{x}}|{\bf{y}}} \right)}}{{q\left( {{\bf{x}}|{\bf{y}}} \right)}} \ge 0}$ gives:
\be\label{eq:v_bound}
I\left( {{\bf{x}}:{\bf{y}}} \right) \ge H\left( {\bf{x}} \right) + {\left\langle {\ln q\left( {{\bf{x}}|{\bf{y}}} \right)} \right\rangle _{p\left( {{\bf{x}},{\bf{y}}} \right)}}
\ee
where angled brackets represent averages and $q(\vx|\vy)$ is an arbitrary variational distribution. This bound becomes exact if $q(\vx|\vy) \equiv p(\vx|\vy)$.

It is worthwhile to note that in the context of unsupervised representation learning, $p(\vy|\vx)$ and $q(\vx|\vy)$ can be viewed as an {\em encoder} and a {\em decoder}, respectively. In this case, $\vy$ needs to be learned by maximizing the lower bound in Eq.~\ref{eq:v_bound}  by iteratively adjusting the parameters of the encoder and decoder, such as~\cite{agakov2004algorithm, mohamed2015variational}.

\subsection{Variational Information Maximization for Feature Selection}

Naturally, in terms of information-theoretic feature selection, we could also try to optimize the variational lower bound in Eq.~\ref{eq:v_bound} by choosing a subset of features $S^*$ in $\vx$, such that,
\be \label{eq:lb_subset_old}
{S^*} = \mathop {\arg \max }\limits_S \left\{ {H\left( {{{\bf{x}}_S}} \right) + {{\left\langle {\ln q\left( {{{\bf{x}}_S}|{\bf{y}}} \right)} \right\rangle }_{p\left( {{{\bf{x}}_S},{\bf{y}}} \right)}}} \right\}
\ee
 However, the $H(\vx_S)$ term in RHS of Eq.~\ref{eq:lb_subset_old} is still intractable when $\vx_S$ is very high-dimensional. 

Nonetheless, by noticing that variable $\vy$ is the class label, which is usually discrete, and hence $H(\vy)$ is fixed and tractable, by symmetry we switch $\vx$ and $\vy$ in Eq.~\ref{eq:v_bound} and rewrite the lower bound as follows:
\be \label{eq:v_bound_fs}
I\left( {{\bf{x}}:{\bf{y}}} \right) &\ge H\left( {\bf{y}} \right) + {\left\langle {\ln q\left( {{\bf{y}}|{\bf{x}}} \right)} \right\rangle _{p\left( {{\bf{x}},{\bf{y}}} \right)}}\\
 &= {\left\langle {\ln \left( {\frac{{q\left( {{\bf{y}}|{\bf{x}}} \right)}}{{p\left( {\bf{y}} \right)}}} \right)} \right\rangle _{p\left( {{\bf{x}},{\bf{y}}} \right)}}
\ee

The equality in Eq.~\ref{eq:v_bound_fs} is obtained by noticing that $H(\vy)={\left\langle { - \ln p\left( {\bf{y}} \right)} \right\rangle _{p\left( {\bf{y}} \right)}}$.

By using Eq.~\ref{eq:v_bound_fs}, the lower bound optimal subset $S^*$ of $\vx$ becomes:
\be \label{eq:lb_subset_new}
{S^*} = \mathop {\arg \max }\limits_S \left\{ {{{\left\langle {\ln \left( {\frac{{q\left( {{\bf{y}}|{{\bf{x}}_S}} \right)}}{{p\left( {\bf{y}} \right)}}} \right)} \right\rangle }_{p\left( {{{\bf{x}}_S},{\bf{y}}} \right)}}} \right\}
\ee

\subsubsection{Choice of Variational Distribution}

$q(\vy|\vx_S)$ in Eq.~\ref{eq:lb_subset_new} can be \textit{any} distribution as long as it is normalized. We need to choose $q(\vy|\vx_S)$ to be as general as possible while still keeping the term ${\left\langle {\ln q\left( {{\bf{y}}|{\bf{x}}_S} \right)} \right\rangle _{p\left( {{\bf{x}}_S,{\bf{y}}} \right)}}$ tractable in Eq.~\ref{eq:lb_subset_new}. 

As a result, we set $q(\vy|\vx_S)$ as 
\be \label{eq:q_general}
q\left( {{\bf{y}}|{{\bf{x}}_S}} \right) = \frac{{q\left( {{{\bf{x}}_S},{\bf{y}}} \right)}}{{q\left( {{{\bf{x}}_S}} \right)}} = \frac{{q\left( {{{\bf{x}}_S}|{\bf{y}}} \right)p\left( {\bf{y}} \right)}}{{\sum\limits_{{\bf{y}}'} {q\left( {{{\bf{x}}_S}|{\bf{y}}'} \right)p\left( {{\bf{y}}'} \right)} }}
\ee

We can verify that Eq.~\ref{eq:q_general} is normalized even if $q(\vx_S|\vy)$ is not normalized.

If we further denote,
\be \label{eq:qx}
q\left( {{{\bf{x}}_S}} \right) = \sum\limits_{{\bf{y}}'} {q\left( {{{\bf{x}}_S}|{\bf{y}}'} \right)p\left( {{\bf{y}}'} \right)} 
\ee

then by combining Eqs.~\ref{eq:lb_subset_new},~\ref{eq:q_general}, we get,
\be\label{eq:v_bound_abbr}
I\left( {{{\bf{x}}_S}:{\bf{y}}} \right) \ge {\left\langle {\ln \left( {\frac{{q\left( {{{\bf{x}}_S}|{\bf{y}}} \right)}}{{q\left( {{{\bf{x}}_S}} \right)}}} \right)} \right\rangle _{p\left( {{{\bf{x}}_S},{\bf{y}}} \right)}} \equiv {I_{LB}}\left( {{{\bf{x}}_S}:{\bf{y}}} \right)\ee

{\bf Auto-Regressive Decomposition.\quad} 
Now that $q(\vy|\vx_S)$ is defined, all we need to do is model $q(\vx_S|\vy)$ under Eq.~\ref{eq:q_general}, and $q(\vx_S)$ is easy to compute based on $q(\vx_S|\vy)$. Here we decompose $q(\vx_S|\vy)$ as an auto-regressive distribution assuming $T$ features in $S$:
\be\label{eq:v_auto}
q\left( {{{\bf{x}}_S}|{\bf{y}}} \right) = q\left( {{{\bf{x}}_{{f_1}}}|{\bf{y}}} \right)\prod\limits_{t = 2}^T {q\left( {{{\bf{x}}_{{f_t}}}|{{\bf{x}}_{{f_{ < t}}}},{\bf{y}}} \right)}
\ee
\begin{figure}[htbp] 
   \centering
   \includegraphics[scale=0.4]{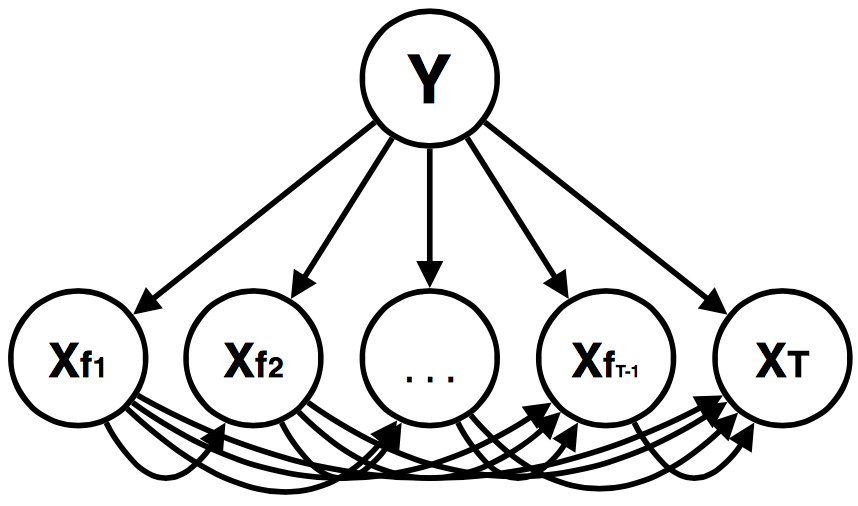} 
   \caption{Auto-regressive decomposition for $q(\vx_S|\vy)$}
   \label{fig:auto}
\end{figure}where $\vx_{f_{<t}}$ denotes $\{\vx_{f_1},\vx_{f_2},...,\vx_{f_{t-1}}\}$. The graphical model in Fig.~\ref{fig:auto} demonstrates this decomposition. The main advantage of this model is that it is well-suited for the forward feature selection procedure where one feature is selected at a time (which we will explain in Sec.~\ref{sec:algo}). And if ${q\left( {{{\bf{x}}_{{f_t}}}|{{\bf{x}}_{{f_{ < t}}}},{\bf{y}}} \right)}$ is tractable, then so is the whole distribution $q(\vx_S|\vy)$. Therefore, we would find tractable $Q$-Distributions over 
${q\left( {{{\bf{x}}_{{f_t}}}|{{\bf{x}}_{{f_{ < t}}}},{\bf{y}}} \right)}$. Below we illustrate two such $Q$-distributions.

{\bf Naive Bayes $Q$-distribution. \quad}
An natural idea would be to assume $\vx_t$ is independent of other variables given $\vy$, i.e., 
\be\label{eq:v_distri}
q\left( {{{\bf{x}}_{{f_t}}}|{{\bf{x}}_{{f_{ < t}}}},{\bf{y}}} \right) = p\left( {{{\bf{x}}_{{f_t}}}|{\bf{y}}} \right)
\ee


Then the variational distribution $q(\vy|\vx_S)$ can be written based on Eqs.~\ref{eq:q_general} and~\ref{eq:v_distri} as follows:
\be\label{eq:v_distri_real}
q\left( {{\bf{y}}|{{\bf{x}}_S}} \right) = \frac{{p\left( {\bf{y}} \right)\prod\limits_{j \in S} {p\left( {{{\bf{x}}_j}|{\bf{y}}} \right)} }}{{\sum\limits_{{{\bf{y}}^\prime }} {p\left( {{{\bf{y}}^\prime }} \right)\prod\limits_{j \in S} {p\left( {{{\bf{x}}_j}|{{\bf{y}}^\prime }} \right)} } }}
\ee

And we also have the following theorem:
\begin{theorem}[Exact Naive Bayes]\label{theo:naive}
Under Eq.~\ref{eq:v_distri_real}, the lower bound in Eq.~\ref{eq:v_bound_fs} becomes exact if and only if data is generated by a Naive Bayes model, i.e., $p\left( {{\bf{x}},{\bf{y}}} \right) = p\left( {\bf{y}} \right)\prod\limits_i {p\left( {{{\bf{x}}_i}|{\bf{y}}} \right)}$.
\end{theorem}
The proof for Theorem~\ref{theo:naive} becomes obvious by using the mutual information definition. Note that the most-cited MI-based feature selection method mRMR~\cite{peng2005feature} also assumes conditional independence given the class label $\vy$ as shown in~\cite{brown2012conditional, balagani2010feature, vinh2015can}, but they make additional stronger independence assumptions among only feature variables. 

{\bf Pairwise $Q$-distribution. \quad} 
We now consider an alternative approach that is more general than the Naive Bayes distribution:
\be\label{eq:v_pairwise_distri}
q\left( {{{\bf{x}}_{{f_t}}}|{{\bf{x}}_{{f_{ < t}}}},{\bf{y}}} \right) = {\left( {\prod\limits_{i = 1}^{t - 1} {p\left( {{{\bf{x}}_{{f_t}}}|{{\bf{x}}_{{f_i}}},{\bf{y}}} \right)} } \right)^{\frac{1}{{t - 1}}}}
\ee
In Eq.~\ref{eq:v_pairwise_distri}, we assume $q\left( {{{\bf{x}}_{{f_t}}}|{{\bf{x}}_{{f_{ < t}}}},{\bf{y}}} \right)$ to be the geometric mean of conditional distributions $q(\vx_{f_t}|\vx_{f_i},\vy)$. This assumption is tractable as well as reasonable because if the data is generated by a Naive Bayes model, the lower bound in Eq.~\ref{eq:v_bound_fs} also becomes exact using Eq.~\ref{eq:v_pairwise_distri} due to $p\left( {{{\bf{x}}_{{f_t}}}|{{\bf{x}}_{{f_i}}},{\bf{y}}} \right) \equiv p\left( {{{\bf{x}}_{{f_t}}}|{\bf{y}}} \right)$ in that case.

\subsubsection{Estimating Lower Bound From Data}

Assuming either Naive Bayes $Q$-distribution or Pairwise $Q$-distribution, it is convenient to estimate $q(\vx_S|\vy)$ and $q(\vx_S)$ in Eq.~\ref{eq:v_bound_abbr} by using plug-in probability estimators for discrete data or one/two-dimensional density estimator for continuous data.  We also use the sample mean to approximate the expectation term in Eq.~\ref{eq:v_bound_abbr}. Our final estimator for ${I_{LB}}\left( {{\bf{x}}_S:{\bf{y}}} \right)$ is written as follows:
\be \label{eq:lb_est}
{{\widehat I}_{LB}}\left( {{{\bf{x}}_S}:{\bf{y}}} \right) = \frac{1}{N}\sum\limits_{{{\bf{x}}^{\left( k \right)}},{{\bf{y}}^{\left( k \right)}}} {\ln \frac{{\widehat q\left( {{\bf{x}}_S^{\left( k \right)}|{{\bf{y}}^{\left( k \right)}}} \right)}}{{\widehat q\left( {{\bf{x}}_S^{\left( k \right)}} \right)}}} 
\ee
where $\left\{ {{{\bf{x}}^{\left( k \right)}},{{\bf{y}}^{\left( k \right)}}} \right\}$ are samples from data, and $\widehat q(\cdot)$ denotes the estimate for $q(\cdot)$.

\subsubsection{Variational Forward Feature Selection Under Auto-Regressive Decomposition}\label{sec:algo}

After defining $q(\vy|\vx_S)$ in Eq.~\ref{eq:q_general} and auto-regressive decomposition of $q(\vx_S|\vy)$ in Eq.~\ref{eq:v_distri}, we are able to do the forward feature selection previously described in Eq.~\ref{eq:fw_sel}, but replace the mutual information with its lower bound $\widehat I_{LB}$. Recall that $S^{t-1}$ is  the set of selected features after step $t-1$, then the feature $f_{t}$ will be selected at step $t$ such that
\be\label{eq:fw_sel_lb}
{f_{t}} = \mathop {\arg \max }\limits_{i \notin {S^{t-1}}}  \widehat I_{LB}\left( {{{\bf{x}}_{{S^{t-1}} \cup i}}:{\bf{y}}} \right)
\ee

where $\widehat I_{LB}\left( {{{\bf{x}}_{{S^{t-1}} \cup i}}:{\bf{y}}} \right)$ can be obtained from $\widehat I_{LB}\left( {{{\bf{x}}_{ {S^{t-1}}}}:{\bf{y}}} \right)$ recursively by auto-regressive decomposition $q\left( {{{\bf{x}}_{{S^{t-1}} \cup i}}|{\bf{y}}} \right) = q\left( {{{\bf{x}}_{{S^{t-1}}}}|{\bf{y}}} \right)q\left( {{{\bf{x}}_i}|{{\bf{x}}_{{S^{t-1}}}},{\bf{y}}} \right)$ where $q\left( {{{\bf{x}}_{{S^{t-1}}}}|{\bf{y}}} \right)$ is stored at step $t-1$.

This forward feature selection can be done under auto-regressive decomposition in Eqs.~\ref{eq:q_general} and~\ref{eq:v_auto} for \textit{any} $Q$-distribution. However, calculating $q(\vx_i|\vx_{S^{t}},\vy)$ may vary according to different $Q$-distributions. We can verify that it is easy to get $q(\vx_i|\vx_{S^{t}},\vy)$ recursively from $q(\vx_i|\vx_{S^{t-1}},\vy)$ under Naive Bayes or Pairwise $Q$-distribution. We call our algorithm under these two $Q$-distributions  $\mathcal{VMI}_{naive}$ and $\mathcal{VMI}_{pairwise}$ respectively. 

It is worthwhile noting that the lower bound does not always increase at each step. A decrease in lower bound at step $t$ indicates that the $Q$-distribution would approximate the underlying distribution worse than it did at previous step $t-1$. In this case, the algorithm would re-maximize the lower bound from zero with only the remaining unselected features. We summarize the concrete implementation of our algorithms in supplementary Sec.~\ref{sec:algor}.

{\bf Time Complexity. \quad}
Although our algorithm needs to calculate the distributions at each step, we only need to calculate the probability value at each sample point. For both $\mathcal{VMI}_{naive}$ and $\mathcal{VMI}_{pairwise}$, the total computational complexity is $O(NDT)$ assuming $N$ as number of samples, $D$ as total number of features, $T$ as number of final selected features. The detailed time analysis is left for the supplementary Sec.~\ref{sec:algor}. As shown in Table~\ref{tb:time}, our methods $\mathcal{VMI}_{naive}$ and $\mathcal{VMI}_{pairwise}$ have the same time complexity as mRMR~\cite{peng2005feature}, while state-of-the-art global optimization method $\mathcal{SPEC_{CMI}}$~\cite{nguyen2014effective} is required to precompute the pairwise mutual information matrix, which gives an time complexity of $O(ND^2)$.

\begin{table}[htbp]
\centering
\caption{\textbf{Time complexity in number of features $D$, selected number of features $d$, and number of samples $N$}}
\begin{tabular}{c|c|c|c|c}\label{tb:time}
\textbf{Method} & mRMR & $\mathcal{VMI}_{naive}$ & $\mathcal{VMI}_{pairwise}$ & $\mathcal{SPEC_{CMI}}$ \\
\hline
\textbf{Complexity} & $O(NDT)$ & $O(NDT)$ & $O(NDT)$ & $O(ND^2)$ \\
\end{tabular}
\end{table}
{\bf Optimality Under Tree Graphical Models. \quad}\label{sec:optimality} 
Although our method $\mathcal{VMI}_{naive}$ assumes a Naive Bayes model, we can prove that this method is still optimal if the data is generated according to tree graphical models. Indeed, both of our methods, $\mathcal{VMI}_{naive}$ and $\mathcal{VMI}_{pairwise}$, will always prioritize the first layer features, as shown in Fig.~\ref{fig:synthetic}. This optimality is summarized in Theorem~\ref{theo:opt} in supplementary Sec.~\ref{sec:optt}.


\section{Experiments}\label{sec:exp}
We begin with the experiments on a synthetic model according to the tree structure illustrated in the left part of Fig.~\ref{fig:synthetic}. The detailed data generating process is shown in supplementary section~\ref{sec:syn}. The root node $\bf Y$ is a binary variable, while other variables are continuous. We use $\mathcal{VMI}_{naive}$ to optimize the lower bound $I_{LB}(\vx:\vy)$. $5000$ samples are used to generate the synthethic data, and variational $Q$-distributions are estimated by kernel density estimator. We can see from the plot in the right part of Fig.~\ref{fig:synthetic} that our algorithm, $\mathcal{VMI}_{naive}$, selects $\vx_1$, $\vx_2$, $\vx_3$ as the first three features, although $\vx_2$ and $\vx_3$ are only weakly correlated with $\vy$. If we continue to add deeper level features $\{\vx_4,...,\vx_9\}$, the lower bound will decrease. For comparison, we also illustrate the mutual information between each single feature $\vx_i$ and $\vy$ in Table~\ref{tb:mi_syn}. We can see from Table~\ref{tb:mi_syn} that it would choose $\vx_1$, $\vx_4$ and $\vx_5$ as the top three features by using the maximum relevance criteria~\cite{lewis1992feature}.
\begin{figure}[htbp] 
\centering
  \raisebox{0.1\height}{\includegraphics[width=0.35\columnwidth]{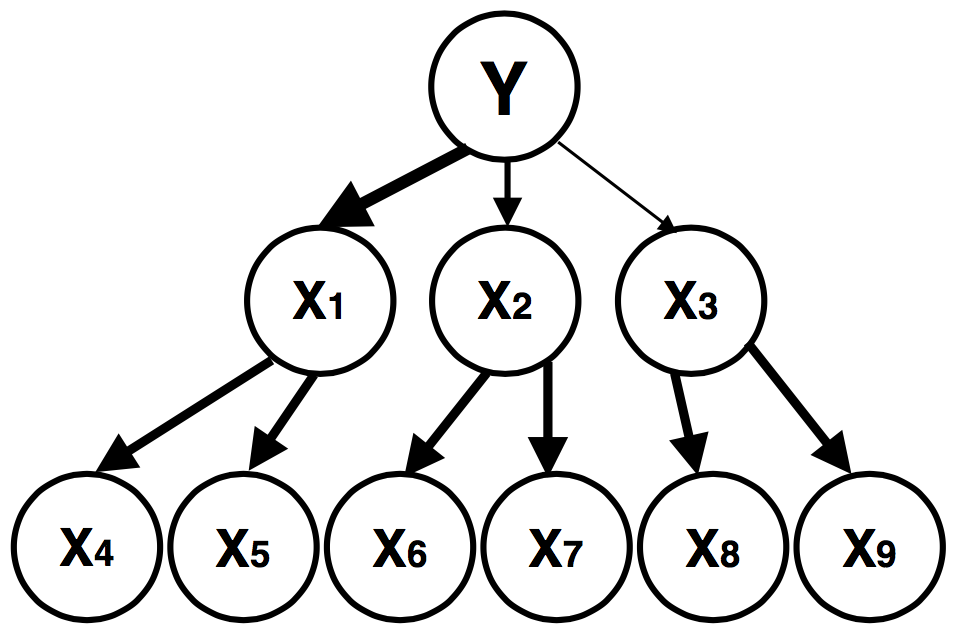}} \quad
     \raisebox{0.0\height}{\includegraphics[width=0.35\columnwidth]{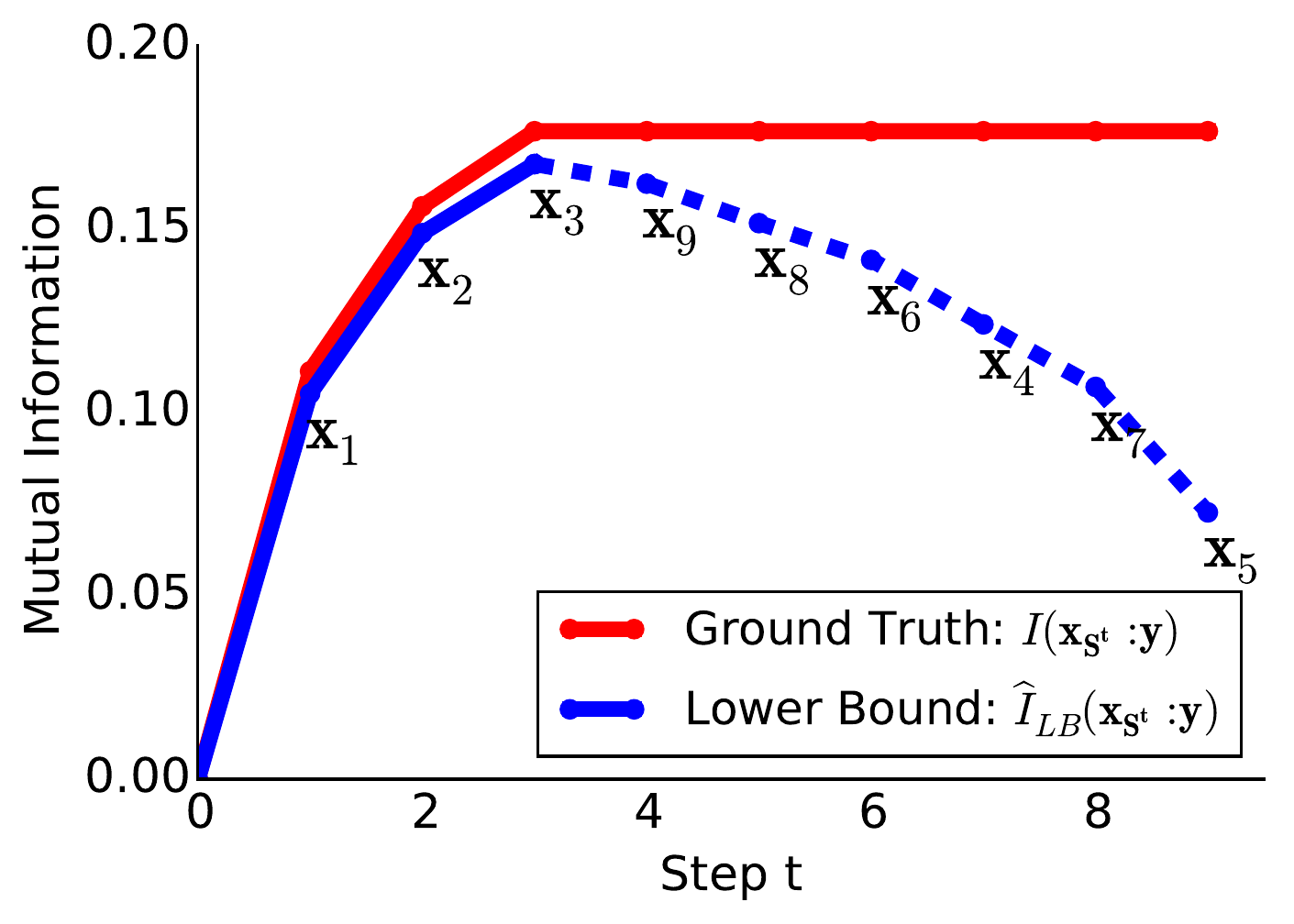}}
\caption{(Left) This is the generative model used for synthetic experiments. Edge thickness represents the relationship strength. (Right) Optimizing the lower bound by $\mathcal{VMI}_{naive}$. Variables under the blue line denote the features selected at each step. Dotted blues line shows the decreasing lower bound if adding more features. Ground-truth mutual information is obtained using $N=100,000$ samples.
}
\label{fig:synthetic}
\end{figure}
\begin{table}[htbp]
\begin{center}
\begin{tabular}{| c | c | c | c | c | c | c | c | c | c |}
  \hline
  feature$_i$ & $\vx_1$ & $\vx_2$ & $\vx_3$ & $\vx_4$ & $\vx_5$  & $\vx_6$ & $\vx_7$ & $\vx_8$ & $\vx_9$ \\
  \hline
$I(\vx_i:\vy)$ & \bf 0.111 & 0.052 & 0.022 & \bf 0.058 & \bf 0.058  & 0.025 & 0.029 & 0.012 & 0.013  \\       
 \hline
\end{tabular}
\end{center}
\caption{Mutual information between label $\vy$ and each feature $\vx_i$ for Fig.~\ref{fig:synthetic}. $I(\vx_i:\vy)$ is estimated using N=100,000 samples. Top three variables with highest mutual information are highlighted in bold.} \label{tb:mi_syn}
\end{table}

\subsection{Real-World Data}
We compare our algorithms $\mathcal{VMI}_{naive}$ and $\mathcal{VMI}_{pairwise}$ with other popular information-theoretic feature selection methods, \shuyang{including mRMR~\cite{peng2005feature}, JMI~\cite{yang1999data}, MIM~\cite{lewis1992feature}, CMIM~\cite{fleuret2004fast}, CIFE~\cite{lin2006conditional}, and $\mathcal{SPEC_{CMI}}$~\cite{nguyen2014effective}.} We use 17 well-known datasets in previous feature selection studies~\cite{brown2012conditional, nguyen2014effective} (all data are discretized). The dataset summaries are illustrated in supplementary Sec.~\ref{sec:data}. We use the average cross-validation error rate on the range of 10 to 100 features to compare different algorithms under the same setting as~\cite{nguyen2014effective}. 10-fold cross-validation is employed for datasets with number of samples $N\ge100$ and leave-one-out cross-validation otherwise. The 3-Nearest-Neighbor classifier is used for Gisette and Madelon, following~\cite{brown2012conditional}. While for the remaining datasets, the classifier is chosen to be Linear SVM, following~\cite{rodriguez2010quadratic,nguyen2014effective}. 

The experimental results can be seen in Table~\ref{tb:results}\footnote{we omit the results for $MIM$ and $CIFE$ due to space limitations, the complete results are shown in the supplementary Sec.~\ref{sec:data}.}. The entries with $*$ and $**$ indicate the best performance and the second best performance respectively (in terms of average error rate). We also use the paired t-test at 5\% significant level to test the hypothesis that $\mathcal{VMI}_{naive}$ or $\mathcal{VMI}_{pairwise}$ performs significantly better than other methods, or vice visa. Overall, we find that both of our methods, $\mathcal VMI_{naive}$ and $\mathcal VMI_{pairwise}$, strongly outperform other methods, indicating our variational feature selection framework is a promising addition to the current literature of information-theoretic feature selection.

\begin{table}[htbp]
\caption{\textbf{Average cross-validation error rate comparison of $\mathcal{VMI}$ against other methods. The last two lines indicate win(W)/tie(T)/loss(L) for $\mathcal{VMI}_{naive}$ and $\mathcal{VMI}_{pairwise}$ respectively.}}
\small
\label{tb:results}
\hspace*{-0.7cm}\begin{tabular}{|ccccc||c|c|}
\hline
Dataset & mRMR & JMI & CMIM & $\mathcal{SPEC_{CMI}}$ & $\mathcal{VMI}_{naive}$ & $\mathcal{VMI}_{pairwise}$ \\
\hline
Lung & \bf 10.9$\pm$(4.7)$^{**}$ & 11.6$\pm$(4.7) & 11.4$\pm$(3.0) & 11.6$\pm$(5.6) & \ \  \bf 7.4$\pm$(3.6)$^*$ & 14.5$\pm$(6.0) \\
Colon & 19.7$\pm$(2.6) & 17.3$\pm$(3.0) & 18.4$\pm$(2.6) & 16.1$\pm$(2.0) & \bf 11.2$\pm$(2.7)$^*$ & \bf 11.9$\pm$(1.7)$^{**}$ \\
Leukemia & \ \  0.4$\pm$(0.7) & \ \  1.4$\pm$(1.2) & \ \  1.1$\pm$(2.0) & \ \  1.8$\pm$(1.3) & \ \  \bf 0.0$\pm$(0.1)$^*$ & \ \  \bf 0.2$\pm$(0.5)$^{**}$ \\
Lymphoma & \ \  5.6$\pm$(2.8) & \ \  6.6$\pm$(2.2) & \ \  8.6$\pm$(3.3) & 12.0$\pm$(6.6) & \ \  \bf 3.7$\pm$(1.9)$^*$ & \ \  \bf 5.2$\pm$(3.1)$^{**}$ \\
Splice & \bf 13.6$\pm$(0.4)$^*$ & \bf 13.7$\pm$(0.5)$^{**}$ & 14.7$\pm$(0.3) & \bf 13.7$\pm$(0.5)$^{**}$ & \bf 13.7$\pm$(0.5)$^{**}$ & \bf 13.7$\pm$(0.5)$^{**}$ \\
Landsat & 19.5$\pm$(1.2) & 18.9$\pm$(1.0) & 19.1$\pm$(1.1) & 21.0$\pm$(3.5) & \bf 18.8$\pm$(0.8)$^*$ & \bf 18.8$\pm$(1.0)$^{**}$ \\
Waveform & \bf 15.9$\pm$(0.5)$^*$ & \bf 15.9$\pm$(0.5)$^*$ & 16.0$\pm$(0.7) & \bf 15.9$\pm$(0.6)$^{**}$ & \bf 15.9$\pm$(0.6)$^{**}$ & \bf 15.9$\pm$(0.5)$^*$ \\
KrVsKp & \ \  \bf 5.1$\pm$(0.7)$^{**}$ & \ \  5.2$\pm$(0.6) & \ \  5.3$\pm$(0.5) & \ \  \bf 5.1$\pm$(0.6)$^{*}$ & \ \  5.3$\pm$(0.5) & \ \  \bf 5.1$\pm$(0.7)$^{**}$ \\
Ionosphere & 12.8$\pm$(0.9) & 16.6$\pm$(1.6) & 13.1$\pm$(0.8) & 16.8$\pm$(1.6) & \bf 12.7$\pm$(1.9)$^{**}$ & \bf 12.0$\pm$(1.0)$^*$ \\
Semeion & 23.4$\pm$(6.5) & 24.8$\pm$(7.6) & 16.3$\pm$(4.4) & 26.0$\pm$(9.3) & \bf 14.0$\pm$(4.0)$^*$ & \bf 14.5$\pm$(3.9)$^{**}$ \\
Multifeat. & \ \  4.0$\pm$(1.6) & \ \  4.0$\pm$(1.6) & \ \  3.6$\pm$(1.2) & \ \  4.8$\pm$(3.0) & \ \  \bf 3.0$\pm$(1.1)$^*$ & \ \  \bf 3.5$\pm$(1.1)$^{**}$ \\
Optdigits & \ \  7.6$\pm$(3.3) & \ \  7.6$\pm$(3.2) & \ \  \bf 7.5$\pm$(3.4)$^{**}$ & \ \  9.2$\pm$(6.0) & \ \  \bf 7.2$\pm$(2.5)$^*$ & \ \  7.6$\pm$(3.6) \\
Musk2 & \bf 12.4$\pm$(0.7)$^*$ & 12.8$\pm$(0.7) & 13.0$\pm$(1.0) & 15.1$\pm$(1.8) & 12.8$\pm$(0.6) & \bf 12.6$\pm$(0.5)$^{**}$ \\
Spambase & \ \  6.9$\pm$(0.7) & \ \  7.0$\pm$(0.8) & \ \  \bf 6.8$\pm$(0.7)$^{**}$ & \ \  9.0$\pm$(2.3) & \ \  \bf 6.6$\pm$(0.3)$^*$ & \ \  \bf 6.6$\pm$(0.3)$^*$ \\
Promoter & 21.5$\pm$(2.8) & 22.4$\pm$(4.0)  & 22.1$\pm$(2.9) & 24.0$\pm$(3.7) & \bf 21.2$\pm$(3.9)$^{**}$ & \bf 20.4$\pm$(3.1)$^*$ \\
Gisette & \ \  5.5$\pm$(0.9) & \ \  5.9$\pm$(0.7) & \ \  5.1$\pm$(1.3) & \ \  7.1$\pm$(1.3) & \ \  \bf 4.8$\pm$(0.9)$^{**}$ & \ \  \bf 4.2$\pm$(0.8)$^*$ \\
Madelon & 30.8$\pm$(3.8) & \bf 15.3$\pm$(2.6)$^{*}$ & 17.4$\pm$(2.6) & \bf 15.9$\pm$(2.5)$^{**}$ & 16.7$\pm$(2.7) & 16.6$\pm$(2.9) \\
\hline
\#$W_1/T_1/L_1$: & 11/4/2 & 10/6/1 & 10/7/0 & 13/2/2 & & \\
\hline
\#$W_2/T_2/L_2$: & 9/6/2 & 9/6/2 & 13/3/1 & 12/3/2 & &\\
\hline
\end{tabular}
\end{table}
\begin{figure}[htbp] 
   \centering 
   \includegraphics[width=0.40\columnwidth]{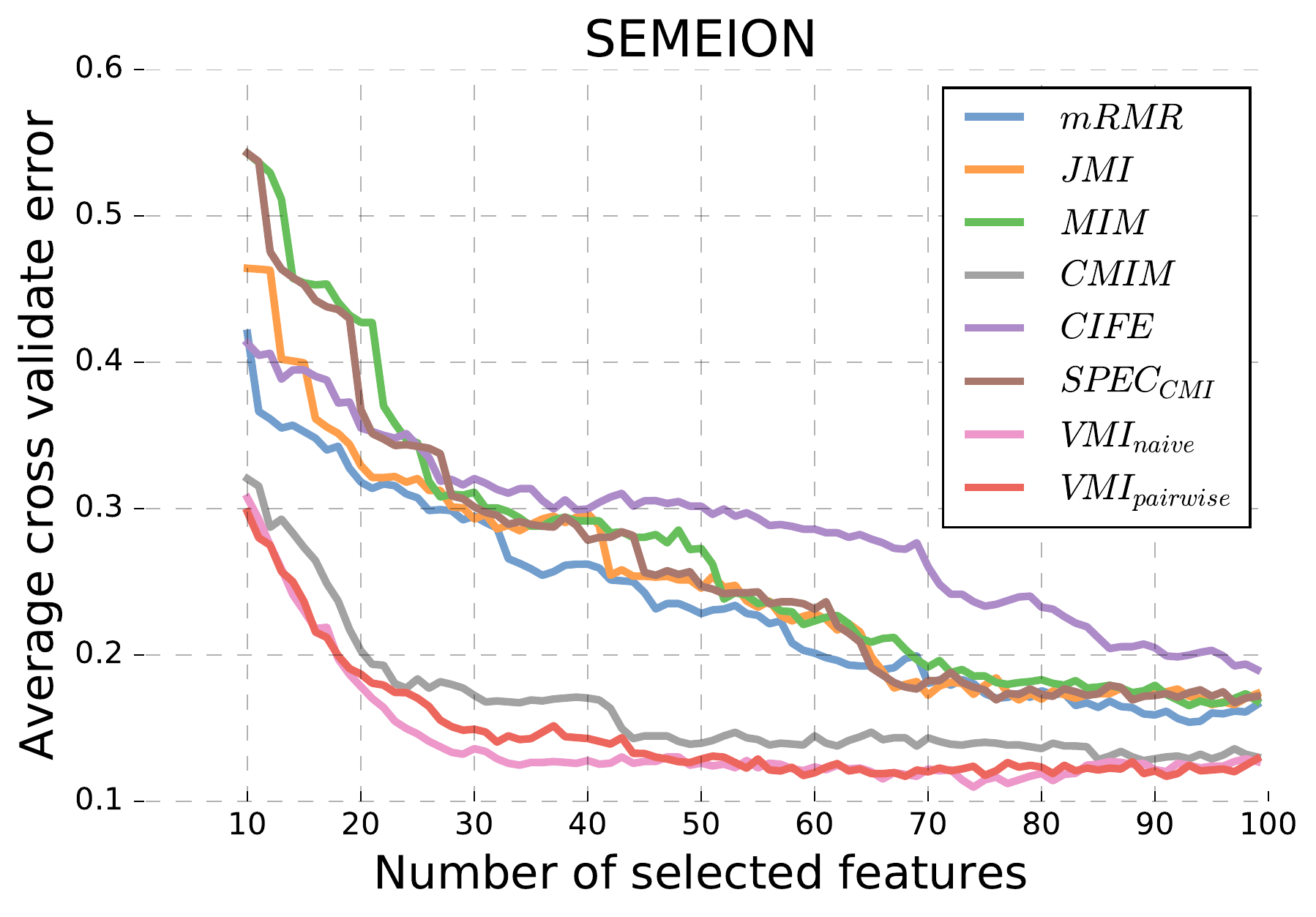} \quad\quad\quad
   \includegraphics[width=0.40\columnwidth]{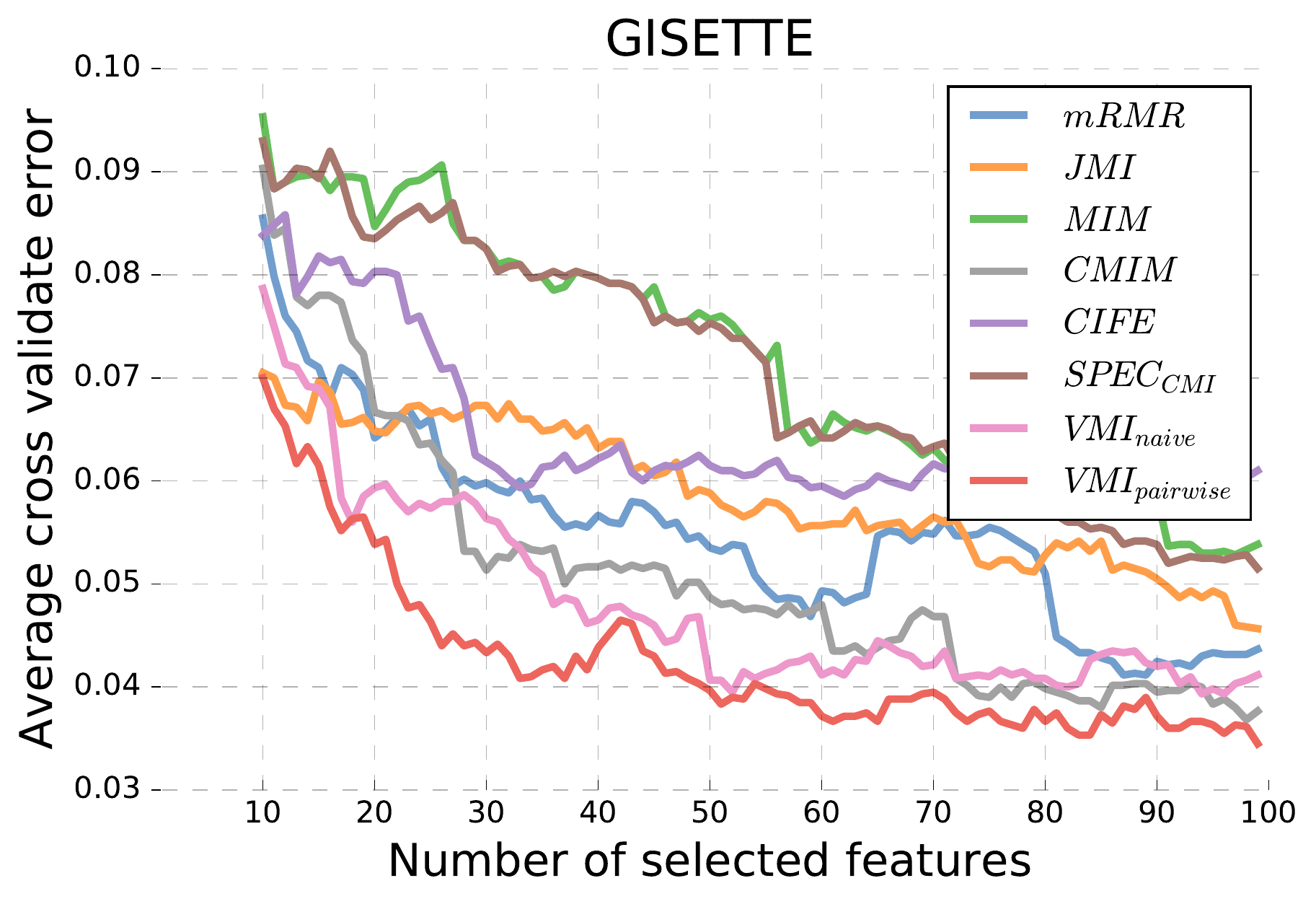} 
   \caption{Number of selected features versus average cross- validation error in datasets Semeion and Gisette.
   }
   \label{fig:real_world} 
\end{figure}

We also plot the average cross- validation error with respect to number of selected features. Fig.~\ref{fig:real_world} shows the two most distinguishable data sets, Semeion and Gisette. We can see that both of our methods, $\mathcal{VMI}_{Naive}$ and $\mathcal{VMI}_{Pairwise}$, have lower error rates in these two data sets.

\section{Related Work}
There has been a significant amount of work on information-theoretic feature selection in the past twenty years:~\cite{brown2012conditional, battiti1994using, yang1999data, fleuret2004fast, peng2005feature, lewis1992feature, rodriguez2010quadratic,nguyen2014effective,cheng2011conditional}, to name a few. Most of these methods are based on combinations of so-called \textit{relevant, redundant} and \textit{complimentary} information. Such combinations representing low-order approximations of mutual information are derived from two assumptions, and it has proved unrealistic to expect both assumptions to be true. Inspired by group testing~\cite{zhou2014parallel}, more scalable feature selection methods have been developed, but this method also requires the calculation of high-dimensional mutual information as a basic scoring function.


Estimating mutual information from data requires an large number of observations---especially when the dimensionality is high. The proposed variational lower bound can be viewed as a way of estimating mutual information between a high-dimensional continuous variable and a discrete variable. Only a few examples exist in  literature~\cite{ross2014mutual} under this setting. We hope our method will shed light on new ways to estimate mutual information, similar to estimating divergences in~\cite{nguyen2010estimating}.

\section{Conclusion}

Feature selection has been a significant endeavor over the past decade. Mutual information gives a general basis for quantifying the informativeness of features. Despite the clarity of mutual information, estimating it can be difficult. While a large number of information-theoretic methods exist, they are rather limited and rely on mutually inconsistent assumptions about underlying data distributions. We introduced a unifying variational mutual information lower bound to address these issues. We showed that by auto-regressive decomposition, feature selection can be done in a forward manner by progressively maximizing the lower bound. We also presented two concrete methods using Naive Bayes and Pairwise $Q$-distributions, which strongly outperform the existing methods. $\mathcal{VMI}_{naive}$ only assumes a Naive Bayes model, but even this simple model outperforms the existing information-theoretic methods, indicating the effectiveness of our variational information maximization framework. We hope that our framework will inspire new mathematically rigorous algorithms for information-theoretic feature selection, such as optimizing the variational lower bound globally and developing more powerful variational approaches for capturing complex dependencies.

{\small
\setlength{\bibsep}{0.2ex plus 0.3ex}
\bibliographystyle{unsrt}

}
\appendix

\clearpage
\counterwithin{figure}{section}

\section*{Supplementary Material for ``Variational Information Maximization for Feature Selection''}
\section{Detailed Algorithm for Variational Forward Feature Selection}\label{sec:algor}

We describe the detailed algorithm for our approach. We also provide open source code implementing $\mathcal{VMI}_{naive}$ and $\mathcal{VMI}_{pairwise}$~\cite{code_anon}.

Concretely, let us suppose class label $\vy$ is discrete and has $L$ different values $\{y_1,y_2,...,y_L\}$; then we define the distribution $q(\vx_{S^t}|\vy)$ vector $Q_t^{\left( k \right)}$ of size $L$ for each sample $\left( {{{\bf{x}}^{\left( k \right)}},{{\bf{y}}^{\left( k \right)}}} \right)$ at step $t$:
\be \label{eq:p_vector}
Q_t^{\left( k \right)} = {\left[ {\widehat q\left( {{\bf{x}}_{{S^t}}^{\left( k \right)}|{\bf{y}} = {y_1}} \right),...,\widehat q\left( {{\bf{x}}_{{S^t}}^{\left( k \right)}|{\bf{y}} = {y_L}} \right)} \right]^T}
\ee

where ${{\bf{x}}_{{S^t}}^{\left( k \right)}}$ denotes the sample $\vx^{(k)}$ projects onto the $\vx_{S^t}$ feature space.

Also, 
We further denote Y of size $L \times 1$ as the distribution vector of $\vy$ as follows:
\be \label{eq:y_vector}
Y = {\left[ {\widehat p\left( {{\bf{y}} = {y_1}} \right),\widehat p\left( {{\bf{y}} = {y_2}} \right),...,\widehat p\left( {{\bf{y}} = {y_L}} \right)} \right]^T}
\ee

Then we are able to rewrite $q(\vx_{S^{t-1}})$ and $q(\vx_{S^{t-1}}|\vy)$ in terms of $Q_{t-1}^{(k)}, Y$ and substitute them into $\widehat I_{LB}(\vx_{S^{t-1}}:\vy)$. 

To illustrate, at step $t-1$ we have, 
\be\label{eq:lb_recur}
{{\widehat I}_{LB}}\left( {{{\bf{x}}_{{S^{t-1}}}}:{\bf{y}}} \right){\rm{ }} = \frac{1}{N}\sum\limits_{{{\bf{x}}^{\left( k \right)}},{{\bf{y}}^{\left( k \right)}}} \log\left( {p\left( {{\bf{x}}_{{S^{t-1}}}^{\left( k \right)}|{\bf{y}} = {{\bf{y}}^{\left( k \right)}}} \right)}\right)  - \frac{1}{N}\sum\limits_k {\log \left( {{Y^T}Q_{t-1}^{\left( k \right)}} \right)} 
\ee

To select a feature $i$ at step $t$, let us define the conditional distribution vector $C_{i,t-1}^{\left( k \right)}$ for each feature $i \notin S^{t-1}$ and each sample $\left( {{{\bf{x}}^{\left( k \right)}},{{\bf{y}}^{\left( k \right)}}} \right)$, i.e., 
\be
C_{i,t-1}^{\left( k \right)} = {\left[ {q\left( {{\bf{x}}_i^{\left( k \right)}|{\bf{x}}_{{S^{t-1}}}^{\left( k \right)},{\bf{y}} = {y_1}} \right),...,q\left( {{\bf{x}}_i^{\left( k \right)}|{\bf{x}}_{{S^{t-1}}}^{\left( k \right)},{\bf{y}} = {y_L}} \right)} \right]^T}
\ee

At step $t$, we use $C_{i,t-1}^{\left( k \right)}$ and $Q_{t-1}^{\left( k \right)}$ previously stored and get,
\be\label{eq:fs_update}
{\widehat I_{LB}}\left( {{{\bf{x}}_{{S^{t-1}} \cup i}}:{\bf{y}}} \right){\rm{ }} &= \frac{1}{N}\sum\limits_{{{\bf{x}}^{\left( k \right)}},{{\bf{y}}^{\left( k \right)}}} \log \left({p\left( {{\bf{x}}_{{S^{t-1}}}^{\left( k \right)}|{\bf{y}} = {{\bf{y}}^{\left( k \right)}}} \right)p\left( {{\bf{x}}_i^{\left( k \right)}|{\bf{x}}_{{S^{t-1}}}^{\left( k \right)},{\bf{y}} = {{\bf{y}}^{\left( k \right)}}} \right)} \right) \\
&~~~~- \frac{1}{N}\sum\limits_k {\log \left( {{Y^T}diag\left( {Q_{t-1}^{\left( k \right)}} \right)C_{i,t-1}^{\left( k \right)}} \right)} 
\ee

We summarize our detailed implementation in Algorithm 1.

\begin{algorithm}[htbp]
\begin{algorithmic}
 \STATE {\bf Data:} {$\left( {{{\bf{x}}^{\left( 1 \right)}},{{\bf{y}}^{\left( 1 \right)}}} \right),\left( {{{\bf{x}}^{\left( 2 \right)}},{{\bf{y}}^{\left( 2 \right)}}} \right),...,\left( {{{\bf{x}}^{\left( N \right)}},{{\bf{y}}^{\left( N \right)}}} \right)$}
 \STATE {\bf Input:} {$T \leftarrow$ \{number of features to select\}}
 \STATE {\bf Output:} {$F \leftarrow$ \{final selected feature set\}}
 \STATE $F \leftarrow \left\{ \varnothing  \right\}$; $S^0 \leftarrow \left\{ \varnothing  \right\}$; $t\leftarrow 1$
 \STATE Initialize $Q_0^{(k)}$ and $C_{i,0}^{(k)}$ for any feature $i$; calculate $Y$
 \WHILE{$|F| < T$}
\STATE ${{\widehat I}_{LB}}\left( {{{\bf{x}}_{{S^{t-1}} \cup i}}:{\bf{y}}} \right) \leftarrow$ \{Eq.~\ref{eq:fs_update} for each $i$ not in $F$\}
 \STATE ${f_{t}} \leftarrow \mathop {\arg \max }\limits_{i \notin {S^{t-1}}} \widehat I_{LB}\left( {{{\bf{x}}_{i \cup {S^{t-1}}}}:{\bf{y}}} \right)$ 
\IF {${\widehat I_{LB}}\left( {{{\bf{x}}_{{S^{t - 1}} \cup {f_t}}}:{\bf{y}}} \right) \le {\widehat I_{LB}}\left( {{{\bf{x}}_{{S^{t - 1}}}}:{\bf{y}}} \right)$}
\STATE Clear $S$; Set $t \leftarrow 1$
\ELSE 

\STATE $F \leftarrow F \cup {f_{t}}$
 \STATE ${S^{t}} \leftarrow {S^{t-1}} \cup {f_{t}}$
 \STATE Update $Q_{t}^{(k)}$ and $C_{i,t}^{(k)}$
 \STATE $t \leftarrow t+1$
\ENDIF
 
\ENDWHILE
\end{algorithmic}
 \caption{Variational Forward Feature Selection (VMI)}

\end{algorithm}
Updating $Q_{t}^{\left( k \right)}$ and $C_{i,t}^{\left( k \right)}$ in Algorithm 1 may vary according to different $Q$-distributions. But we can verify that under Naive Bayes $Q$-distribution or Pairwise $Q$-distribution, $Q_{t}^{\left( k \right)}$ and $C_{i,t}^{\left( k \right)}$ can be obtained recursively from $Q_{t-1}^{\left( k \right)}$ and $C_{i,t-1}^{\left( k \right)}$ by noticing that $q\left( {{{\bf{x}}_i}|{{\bf{x}}_{{S^{t}}}},{\bf{y}}} \right) = p\left( {{{\bf{x}}_i}|{\bf{y}}} \right)$ for Naive Bayes $Q$-distribution and $q\left( {{{\bf{x}}_i}|{{\bf{x}}_{{S^{t}}}},{\bf{y}}} \right) = {\left( {p\left( {{{\bf{x}}_i}|{{\bf{x}}_{{f_{t}}}},y} \right)q{{\left( {{{\bf{x}}_i}|{{\bf{x}}_{{S^{t-1}}}},{\bf{y}}} \right)}^{t-1}}} \right)^{t}}$ for Pairwise $Q$-distribution. 

Let us denote $N$ as number of samples, $D$ as total number of features, $T$ as number of selected features and $L$ as number of distinct values in class variable $\vy$. The computational complexity of Algorithm 1 involves calculating the lower bound for each feature $i$ at every step which is $O(NDL)$; updating $C_{i,t}^{(k)}$ would cost $O(NDL)$ for pairwise $Q$-distribution and $O(1)$ for Naive Bayes $Q$-distribution; updating $Q_t^{(k)}$ would cost $O(NDL)$. We need to select $T$ features, therefore the time complexity is $O(NDT)$\footnote{we ignore $L$ here because the number of classes is usually much smaller.}. 

\section{Optimality under Tree Graphical Models}\label{sec:optt}
\begin{figure}[htbp] 
   \centering
   \includegraphics[scale=0.3]{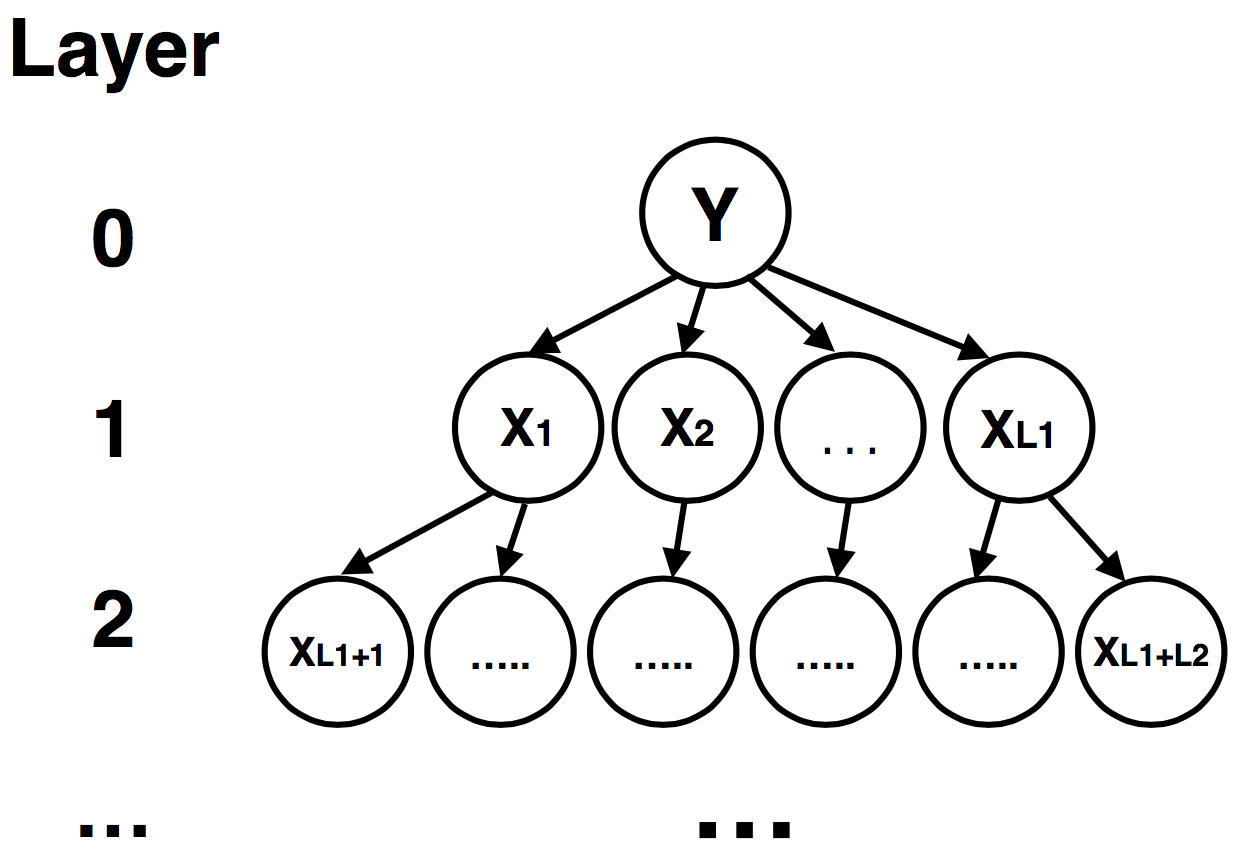} 
   \caption{Demonstration of tree graphical model, label $\vy$ is the root node.}
   \label{fig:tree}
\end{figure}

\begin{theorem}[Optimal Feature Selection]\label{theo:opt}
If data is generated according to tree graphical models, where the class label $\vy$ is the root node, denote the child nodes set in the first layer as $\mathcal L_1 = \{\vx_1, \vx_2, ..., \vx_{L_1}\}$, as shown in Fig.~\ref{fig:tree}. Then there must exist a step $T > 0$ such that the following three conditions hold by using $\mathcal{VMI}_{naive}$ or $\mathcal{VMI}_{pairwise}$:

Condition I: The selected feature set $S^T \subset \mathcal L_1$.

Condition II: $I_{LB}(\vx_{S^t}:\vy)=I(\vx_{S^t}:\vy)$ for $1 \le t \le T$.

Condition III: $I_{LB}(\vx_{S^T}:\vy)=I(\vx:\vy)$.
\end{theorem}

\begin{proof}
 We prove this theorem by induction. For tree graphical model when selecting the first layer features, $\mathcal{VMI}_{naive}$ and $\mathcal{VMI}_{pairwise}$ are mathematically equal, therefore we only prove $\mathcal{VMI}_{naive}$ case and $\mathcal{VMI}_{pairwise}$ follows the same proof.
 
1) At step $t=1$, for each feature $i$, we have,
\be
{I_{LB}}\left( {{{\bf{x}}_i}:{\bf{y}}} \right) &= {\left\langle {\ln \left( {\frac{{q\left( {{{\bf{x}}_i}|{\bf{y}}} \right)}}{{q\left( {{{\bf{x}}_i}} \right)}}} \right)} \right\rangle _{p\left( {{\bf{x}},{\bf{y}}} \right)}} \\
&= {\left\langle {\ln \left( {\frac{{p\left( {{{\bf{x}}_i}|{\bf{y}}} \right)}}{{\sum\limits_{{\bf{y}}'} {p\left( {{\bf{y}}'} \right)p\left( {{{\bf{x}}_i}|{\bf{y}}'} \right)} }}} \right)} \right\rangle _{p\left( {{\bf{x}},{\bf{y}}} \right)}}\\
 &= {\left\langle {\ln \left( {\frac{{p\left( {{{\bf{x}}_i}|{\bf{y}}} \right)}}{{p\left( {{{\bf{x}}_i}} \right)}}} \right)} \right\rangle _{p\left( {{\bf{x}},{\bf{y}}} \right)}} = I\left( {{{\bf{x}}_i}:{\bf{y}}} \right)
\ee
Thus, we are choosing a feature that has the maximum mutual information with $\vy$ at the very first step. Based on the data processing inequality, we have $I(\vx_i:\vy) \ge I(desc(\vx_i):\vy)$ for any $\vx_i$ in layer 1 where $desc(\vx_i)$ represents any descendant of $\vx_i$. Thus, we always select features among the nodes of the first layer at step $t=1$ without loss of generality. If node $\vx_j$ that is not in the first layer is selected at step $t=1$, denote $ances(\vx_j)$ as $\vx_j$'s ancestor in layer 1, then $I(\vx_j:\vy) = I(ances(\vx_j):\vy)$ which means that the information is not lost from $ances(\vx_j) \to \vx_j$. In this case, one can always switch $ances(\vx_j)$ with $\vx_j$ and let $\vx_j$ be in the first layer, which does not conflict with the model assumption.

Therefore, condition I and II are satisfied in step $t=1$.

2) Assuming condition I and II are satisfied in step $t$, then we have the following argument in step $t+1$:

We discuss the candidate nodes in three classes, and argue that nodes in \textit{\textbf{Remaining-Layer 1 Class}} are always being selected.

\textbf{\textit{Redundant Class}} For any descendant $desc(S^t)$ of selected feature set $S^t$, we have,
\be\label{eq:proof_eq1}
I\left( {{{\bf{x}}_{{S^t} \cup desc\left( {{S^t}} \right)}}:{\bf{y}}} \right) = I\left( {{{\bf{x}}_{{S^t}}}:{\bf{y}}} \right) = {I_{LB}}\left( {{{\bf{x}}_{{S^t}}}:{\bf{y}}} \right) 
\ee
Eq.~\ref{eq:proof_eq1} comes from the fact that the $desc(S^t)$ carries no additional information about $\vy$ other than $S^t$. The second equality is by induction.

Based on Eq.~\ref{eq:v_bound_abbr} and~\ref{eq:proof_eq1}, we have,
\be \label{eq:proof_eq2}
{I_{LB}}\left( {{{\bf{x}}_{{S^t} \cup desc\left( {{S^t}} \right)}}:{\bf{y}}} \right) &< I\left( {{{\bf{x}}_{{S^t} \cup desc\left( {{S^t}} \right)}}:{\bf{y}}} \right) \\
&~~~~= I\left( {{{\bf{x}}_{{S^t}}}:{\bf{y}}} \right) \\
\ee
We assume here that the LHS is \textit{strictly} less than RHS in Eq.~\ref{eq:proof_eq2} without loss of generality. This is because if the equality holds, we have $p\left( {{{\rm{x}}_{{S^t}}}|{\bf{y}}} \right)p\left( {desc\left( {{S^t}} \right)|{\bf{y}}} \right) = p\left( {{{\bf{x}}^t},desc\left( {{S^t}} \right)|{\bf{y}}} \right)$ due to Theorem~\ref{theo:naive}. In this case, we can always rearrange $desc(S^t)$ to the first layer, which does not conflict with the model assumption.

Note that by combining Eqs.~\ref{eq:proof_eq1} and ~\ref{eq:proof_eq2}, we can also get 
\be\label{eq:red}
{I_{LB}}\left( {{{\bf{x}}_{{S^t} \cup desc\left( {{S^t}} \right)}}:{\bf{y}}} \right) < {I_{LB}}\left( {{{\bf{x}}_{{S^t}}}:{\bf{y}}} \right)
\ee
Eq.~\ref{eq:red} means that adding a feature in \textit{{Redundant Class}} will actually \textit{decrease} the value of lower bound $I_{LB}$.

\textbf{\textit{Remaining-Layer1 Class}}  For any other unselected node $j$ of the first layer, i.e., $j \in {\mathcal L_1}\backslash {S^t}$, we have
\be\label{eq:proof_eq3}
I\left( {{{\bf{x}}_{{S^t}}}:{\bf{y}}} \right) \le I\left( {{{\bf{x}}_{{S^t} \cup j}}:{\bf{y}}} \right) = {I_{LB}}\left( {{{\bf{x}}_{{S^t} \cup j}}:{\bf{y}}} \right)
\ee
The inequality in Eq.~\ref{eq:proof_eq3} is obvious which comes from the data processing inequality~\cite{cover2012elements}. And the equality in Eq.~\ref{eq:proof_eq3} comes directly from Theorem~\ref{theo:naive}.

\textbf{\textit{Descendants-of-Remaining-Layer1 Class}} For any node $desc(j)$ that is the descendant of $j$ where $j \in {\mathcal L_1}\backslash {S^t}$, we have,
\be \label{eq:proof_eq4}
{I_{LB}}\left( {{{\bf{x}}_{{S^t} \cup desc\left( j \right)}}:{\bf{y}}} \right) &\le  I\left( {{{\bf{x}}_{{S^t} \cup desc\left( j \right)}}:{\bf{y}}} \right) \\
I\left( {{{\bf{x}}_{{S^t} \cup desc\left( j \right)}}:{\bf{y}}} \right) &\le I\left( {{{\bf{x}}_{{S^t} \cup j}}:{\bf{y}}} \right)
\ee
The second inequality of Ineq.~\ref{eq:proof_eq4} also comes from data processing inequality. 

Combining Eqs.~\ref{eq:proof_eq2} and~\ref{eq:proof_eq3}, we get,
\be\label{eq:proof_eq5}
{I_{LB}}\left( {{{\bf{x}}_{{S^t} \cup desc\left( {{S^t}} \right)}}:{\bf{y}}} \right) < {I_{LB}}\left( {{{\bf{x}}_{{S^t} \cup j}}:{\bf{y}}} \right)
\ee

Combining Eqs.~\ref{eq:proof_eq3} and~\ref{eq:proof_eq4}, we get,
\be\label{eq:proof_eq6}
{I_{LB}}\left( {{{\bf{x}}_{{S^t} \cup desc\left( j \right)}}:{\bf{y}}} \right) \le {I_{LB}}\left( {{{\bf{x}}_{{S^t} \cup j}}:{\bf{y}}} \right)
\ee

Ineq.~\ref{eq:proof_eq5} essentially tells us the forward feature selection will always choose \textit{Remaining-Layer1 Class} other than \textit{Redundant Class}.

Ineq.~\ref{eq:proof_eq6} is saying we are choosing \textit{Remaining-Layer1 Class} other than \textit{Descendants-of-Remaining-Layer1 Class} without loss of generality (for the equality concern, we can have the same argument in step $t=1$).

Considering Ineqs.~\ref{eq:proof_eq5} and~\ref{eq:proof_eq6}, in step $t+1$, the algorithm chooses node $j$ in \textit{Remaining-Layer1 Class}, i.e., $j \in {\mathcal L_1}\backslash {S^t}$.

Therefore, condition I and II hold at step $t+1$.
 
At step $t+1$, if ${I_{LB}}\left( {{{\bf{x}}_{{S^t} \cup j}}:{\bf{y}}} \right) = {I_{LB}}\left( {{{\bf{x}}_{{S^t}}}:{\bf{y}}} \right)$ for any $j \in \mathcal L_1\backslash S^t$, that means $I\left( {{{\bf{x}}_{{S^t} \cup j}}:{\bf{y}}} \right) = I\left( {{{\bf{x}}_{{S^t}}}:{\bf{y}}} \right)$. Then we have,
\be~\label{eq:proof_eq7}
I\left( {{{\bf{x}}_{{S^t}}}:{\bf{y}}} \right) = I\left( {{{\bf{x}}_{{\mathcal L_1}}}:{\bf{y}}} \right) = I\left( {{\bf{x}}:{\bf{y}}} \right)
\ee

The first equality in Eq.~\ref{eq:proof_eq7} holds because adding any $j$ in $\mathcal L_1\backslash S^t$ will not increase the mutual information. The second equality is due to the data processing inequality under tree graphical model assumption.

Therefore, if ${I_{LB}}\left( {{{\bf{x}}_{{S^t} \cup j}}:{\bf{y}}} \right) = {I_{LB}}\left( {{{\bf{x}}_{{S^t}}}:{\bf{y}}} \right)$ for any $j \in \mathcal L_1\backslash S^t$, we set $T=t$. Thus by combining condition II and Eq.~\ref{eq:proof_eq7}, we have,
\be
{I_{LB}}\left( {{{\bf{x}}_{{S^T}}}:{\bf{y}}} \right) = I\left( {{{\bf{x}}_{{S^T}}}:{\bf{y}}} \right) = I\left( {{\bf{x}}:{\bf{y}}} \right)
\ee

Then condition III holds. 

\end{proof}
\section{Datasets and Results}\label{sec:data}
Table~\ref{tb:data} summarizes the datasets used in the experiment. Table~5 shows the complete results.
\begin{table}[htbp]

\centering
\caption{\textbf{Dataset summary. $N$: \# samples, $d$: \# features, $L$: \# classes.}}
\label{tb:data}
\begin{tabular}{|ccccc|}
\hline
Data & $N$ & $d$ & $L$ & Source \\
\hline
Lung & 73 & 325 & 20 &~\cite{ding2005minimum}\\
Colon & 62& 2000& 2&~\cite{ding2005minimum}\\
Leukemia & 72& 7070& 2&~\cite{ding2005minimum}\\
Lymphoma & 96& 4026& 9&~\cite{ding2005minimum}\\
Splice & 3175& 60 & 3&~\cite{bache2013uci}\\
Landsat & 6435 & 36& 6&~\cite{bache2013uci}\\
Waveform & 5000& 40& 3&~\cite{bache2013uci}\\
KrVsKp & 3196& 36& 2&~\cite{bache2013uci}\\
Ionosphere & 351& 34& 2&~\cite{bache2013uci}\\
Semeion & 1593& 256& 10&~\cite{bache2013uci}\\
Multifeat. & 2000&649 & 10&~\cite{bache2013uci}\\
Optdigits & 3823& 64& 10&~\cite{bache2013uci}\\
Musk2  & 6598& 166& 2&~\cite{bache2013uci}\\
Spambase & 4601& 57& 2&~\cite{bache2013uci}\\
Promoter & 106& 57& 2&~\cite{bache2013uci}\\
Gisette & 6000& 5000& 2&~\cite{guyon2003introduction}\\
Madelon & 2000& 500& 2&~\cite{guyon2003introduction}\\
\hline
\end{tabular}
\end{table}

\begin{sidewaystable}[htbp]

\label{tb:results_new}
\large
\centering
\begin{tabular}{|ccccccc||c|c|}
\hline
Dataset & mRMR & JMI & MIM & CMIM & CIFE & $\mathcal{SPEC_{CMI}}$ & $\mathcal{VMI}_{naive}$ & $\mathcal{VMI}_{pairwise}$ \\
\hline
Lung & \bf 10.9$\pm$(4.7)$^{**}$ & 11.6$\pm$(4.7) & 18.3$\pm$(5.4) & 11.4$\pm$(3.0) & 23.3$\pm$(5.4) & 11.6$\pm$(5.6) & \ \  \bf 7.4$\pm$(3.6)$^*$ & 14.5$\pm$(6.0) \\
Colon & 19.7$\pm$(2.6) & 17.3$\pm$(3.0) & 22.0$\pm$(4.3) & 18.4$\pm$(2.6) & 23.5$\pm$(4.3) & 16.1$\pm$(2.0) & \bf 11.2$\pm$(2.7)$^*$ & \bf 11.9$\pm$(1.7)$^{**}$ \\
Leukemia & \ \  0.4$\pm$(0.7) & \ \  1.4$\pm$(1.2) & \ \  2.5$\pm$(1.1) & \ \  1.1$\pm$(2.0) & \ \  4.9$\pm$(1.9) & \ \  1.8$\pm$(1.3) & \ \  \bf 0.0$\pm$(0.1)$^*$ & \ \  \bf 0.2$\pm$(0.5)$^{**}$ \\
Lymphoma & \ \  5.6$\pm$(2.8) & \ \  6.6$\pm$(2.2) & 13.0$\pm$(6.4) & \ \  8.6$\pm$(3.3) & 35.6$\pm$(4.3) & 12.0$\pm$(6.6) & \ \  \bf 3.7$\pm$(1.9)$^*$ & \ \  \bf 5.2$\pm$(3.1)$^{**}$ \\
Splice & \bf 13.6$\pm$(0.4)$^*$ & 13.7$\pm$(0.5) & \bf 13.6$\pm$(0.5)$^{**}$ & 13.7$\pm$(0.5) & 14.7$\pm$(0.3) & 13.7$\pm$(0.5) & 13.7$\pm$(0.5) & 13.7$\pm$(0.5) \\
Landsat & 19.5$\pm$(1.2) & 18.9$\pm$(1.0) & 22.0$\pm$(3.8) & 19.1$\pm$(1.1) & 19.7$\pm$(1.7) & 21.0$\pm$(3.5) & \bf 18.8$\pm$(0.8)$^*$ & \bf 18.8$\pm$(1.0)$^{**}$ \\
Waveform & \bf 15.9$\pm$(0.5)$^*$ & \bf 15.9$\pm$(0.5)$^*$ & 16.1$\pm$(0.8) & 16.0$\pm$(0.7) & 22.8$\pm$(2.2) & \bf 15.9$\pm$(0.6)$^{**}$ & \bf 15.9$\pm$(0.6)$^{**}$ & \bf 15.9$\pm$(0.5)$^*$ \\
KrVsKp & \ \ 5.1$\pm$(0.7) & \ \  5.2$\pm$(0.6) & \ \  5.3$\pm$(0.6) & \ \  5.3$\pm$(0.5) & \ \  \bf 5.0$\pm$(0.7)$^*$ & \ \  \bf 5.1$\pm$(0.6)$^{**}$ & \ \  5.3$\pm$(0.5) & \ \ 5.1$\pm$(0.7) \\
Ionosphere & 12.8$\pm$(0.9) & 16.6$\pm$(1.6) & 13.3$\pm$(0.9) & 13.1$\pm$(0.8) & 16.1$\pm$(1.6) & 16.8$\pm$(1.6) & \bf 12.7$\pm$(1.9)$^{**}$ & \bf 12.0$\pm$(1.0)$^*$ \\
Semeion & 23.4$\pm$(6.5) & 24.8$\pm$(7.6) & 26.7$\pm$(9.7) & 16.3$\pm$(4.4) & 28.6$\pm$(5.8) & 26.0$\pm$(9.3) & \bf 14.0$\pm$(4.0)$^*$ & \bf 14.5$\pm$(3.9)$^{**}$ \\
Multifeat. & \ \  4.0$\pm$(1.6) & \ \  4.0$\pm$(1.6) & \ \  4.9$\pm$(2.3) & \ \  3.6$\pm$(1.2) & \ \  7.2$\pm$(3.0) & \ \  4.8$\pm$(3.0) & \ \  \bf 3.0$\pm$(1.1)$^*$ & \ \  \bf 3.5$\pm$(1.1)$^{**}$ \\
Optdigits & \ \  7.6$\pm$(3.3) & \ \  7.6$\pm$(3.2) & \ \  7.9$\pm$(3.9) & \ \  \bf 7.5$\pm$(3.4)$^{**}$ & \ \  8.1$\pm$(4.2) & \ \  9.2$\pm$(6.0) & \ \  \bf 7.2$\pm$(2.5)$^*$ & \ \  7.6$\pm$(3.6) \\
Musk2 & \bf 12.4$\pm$(0.7)$^*$ & 12.8$\pm$(0.7) & 14.0$\pm$(1.2) & 13.0$\pm$(1.0) & 13.2$\pm$(0.6) & 15.1$\pm$(1.8) & 12.8$\pm$(0.6) & \bf 12.6$\pm$(0.5)$^{**}$ \\
Spambase & \ \  6.9$\pm$(0.7) & \ \  7.0$\pm$(0.8) & \ \  7.3$\pm$(0.9) & \ \ \bf 6.8$\pm$(0.7)$^{**}$ & 10.3$\pm$(1.8) & \ \  9.0$\pm$(2.3) & \ \  \bf 6.6$\pm$(0.3)$^*$ & \ \  \bf 6.6$\pm$(0.3)$^*$ \\
Promoter & 21.5$\pm$(2.8) & 22.4$\pm$(4.0) & 21.7$\pm$(3.1) & 22.1$\pm$(2.9) & 27.4$\pm$(3.2) & 24.0$\pm$(3.7) & \bf 21.2$\pm$(3.9)$^{**}$ & \bf 20.4$\pm$(3.1)$^*$ \\
Gisette & \ \  5.5$\pm$(0.9) & \ \  5.9$\pm$(0.7) & \ \  7.2$\pm$(1.2) & \ \  5.1$\pm$(1.3) & \ \  6.5$\pm$(0.8) & \ \  7.1$\pm$(1.3) & \ \  \bf 4.8$\pm$(0.9)$^{**}$ & \ \  \bf 4.2$\pm$(0.8)$^*$ \\
Madelon & 30.8$\pm$(3.8) & \bf 15.3$\pm$(2.6)$^{**}$ & 16.8$\pm$(2.7) & 17.4$\pm$(2.6) & \bf 15.1$\pm$(2.7)$^*$ & 15.9$\pm$(2.5) & 16.7$\pm$(2.7) & 16.6$\pm$(2.9) \\
\hline
\#$W_1/T_1/L_1$: & 11/4/2 & 10/6/1 & 11/6/0 & 10/7/0 & 15/0/2 & 13/2/2 & & \\
\hline
\#$W_2/T_2/L_2$: & 9/6/2 & 9/6/2 & 15/2/0 & 13/3/1 & 15/1/1 & 12/3/2 & &\\
\hline
\end{tabular}
\caption{\textbf{Average cross validation error rate comparison of $\mathcal{VMI}$ against other methods. The last two lines indicate win(W)/tie(T)/loss(L) for $\mathcal{VMI}_{naive}$ and $\mathcal{VMI}_{pairwise}$ respectively.}}
\end{sidewaystable}

\section{Generating Synthetic Data} \label{sec:syn}
Here is a detailed generating process for synthetic tree graphical model data in the experiment.

Draw $\vy \sim Bernoulli(0.5)$

Draw $\vx_1 \sim Gaussian(\sigma=1.0,\mu=\vy)$ 

Draw $\vx_2 \sim Gaussian(\sigma=1.0,\mu=\vy/1.5)$ 

Draw $\vx_3 \sim Gaussian(\sigma=1.0,\mu=\vy/2.25)$ 

Draw $\vx_4 \sim Gaussian(\sigma=1.0,\mu=\vx_1)$

Draw $\vx_5 \sim Gaussian(\sigma=1.0,\mu=\vx_1)$

Draw $\vx_6 \sim Gaussian(\sigma=1.0,\mu=\vx_2)$

Draw $\vx_7 \sim Gaussian(\sigma=1.0,\mu=\vx_2)$

Draw $\vx_8 \sim Gaussian(\sigma=1.0,\mu=\vx_3)$

Draw $\vx_9 \sim Gaussian(\sigma=1.0,\mu=\vx_3)$

\end{document}